\documentclass[conference]{IEEEtran}
\IEEEoverridecommandlockouts

\usepackage{cite}
\usepackage{amsmath,amssymb,amsfonts}
\usepackage{algorithmic}
\usepackage{graphicx}
\usepackage{textcomp}
\usepackage{xcolor}
\usepackage{booktabs}
\def\BibTeX{{\rm B\kern-.05em{\sc i\kern-.025em b}\kern-.08em
    T\kern-.1667em\lower.7ex\hbox{E}\kern-.125emX}}
\begin{document}

\def\emb{{\mathbf{z}}}

\title{DEGSTalk: Decomposed Per-Embedding Gaussian Fields for Hair-Preserving Talking Face Synthesis\\
\thanks{$^*$ Corresponding author

This work was supported by the National Natural Science Foundation of China under Grants 82261138629 and 62320106007;  Guangdong Basic and Applied Basic Research Foundation under Grant 2023A1515010688 and Guangdong Provincial Key Laboratory under Grant 2023B1212060076.}
}

\newcommand{\mysmall}[1]{\scriptsize{\color{gray}{#1}}}
\author{
    \IEEEauthorblockN{
        Kaijun Deng$^{1,2}$, Dezhi Zheng$^{1,2}$, Jindong Xie$^{1,2}$, Jinbao Wang$^{2,3}$, Weicheng Xie$^{1,2}$, Linlin Shen$^{*,1,2,3}$, Siyang Song$^{4}$
    }
    \IEEEauthorblockA{
        $^1$ Computer Vision Institute, School of Computer Science and Software Engineering, Shenzhen University \\
        $^2$ National Engineering Laboratory for Big Data System Computing Technology, Shenzhen University \\
        $^3$ Guangdong Provincial Key Laboratory of Intelligent Information Processing \\
        $^4$ Department of Computer Science, University of Exeter
    }
    \vspace{0.1cm}
    \IEEEauthorblockA{
        \{dengkaijun2023, 2310273055, xiejindong2022\}@email.szu.edu.cn, \{wangjb, wcxie, llshen\}@szu.edu.cn, ss2796@cam.ac.uk
    }
}

\maketitle
\begin{abstract}

Accurately synthesizing talking face videos and capturing fine facial features for individuals with long hair presents a significant challenge. To tackle these challenges in existing methods, we propose a decomposed per-embedding Gaussian fields (DEGSTalk), a 3D Gaussian Splatting (3DGS)-based talking face synthesis method for generating realistic talking faces with long hairs. Our DEGSTalk employs Deformable Pre-Embedding Gaussian Fields, which dynamically adjust pre-embedding Gaussian primitives using implicit expression coefficients. This enables precise capture of dynamic facial regions and subtle expressions. Additionally, we propose a Dynamic Hair-Preserving Portrait Rendering technique to enhance the realism of long hair motions in the synthesized videos. Results show that DEGSTalk achieves improved realism and synthesis quality compared to existing approaches, particularly in handling complex facial dynamics and hair preservation.
Our code will be publicly available at https://github.com/CVI-SZU/DEGSTalk.

\end{abstract}

\begin{IEEEkeywords} 3D Gaussian Splatting, Talking Face Synthesis, Pre-Embedding, Hair-Preserving\end{IEEEkeywords}

\section{Introduction}
\label{sec:intro}

\noindent Talking face synthesis has gained significant attention due to its wide range of applications, including Virtual/Augmented reality (VR/AR), filmmaking, human-computer interaction and virtual agent's facial reaction synthesis \cite{morishima1998real,tanaka2022acceptability,tarasuik2013seeing,song2022learning,song2023react2023}. This technology focuses on generating realistic, animated human faces driven by audio or other input signals, offering exciting possibilities across various industries.

\begin{figure*}[h]
  \centering
  \includegraphics[width=\linewidth]{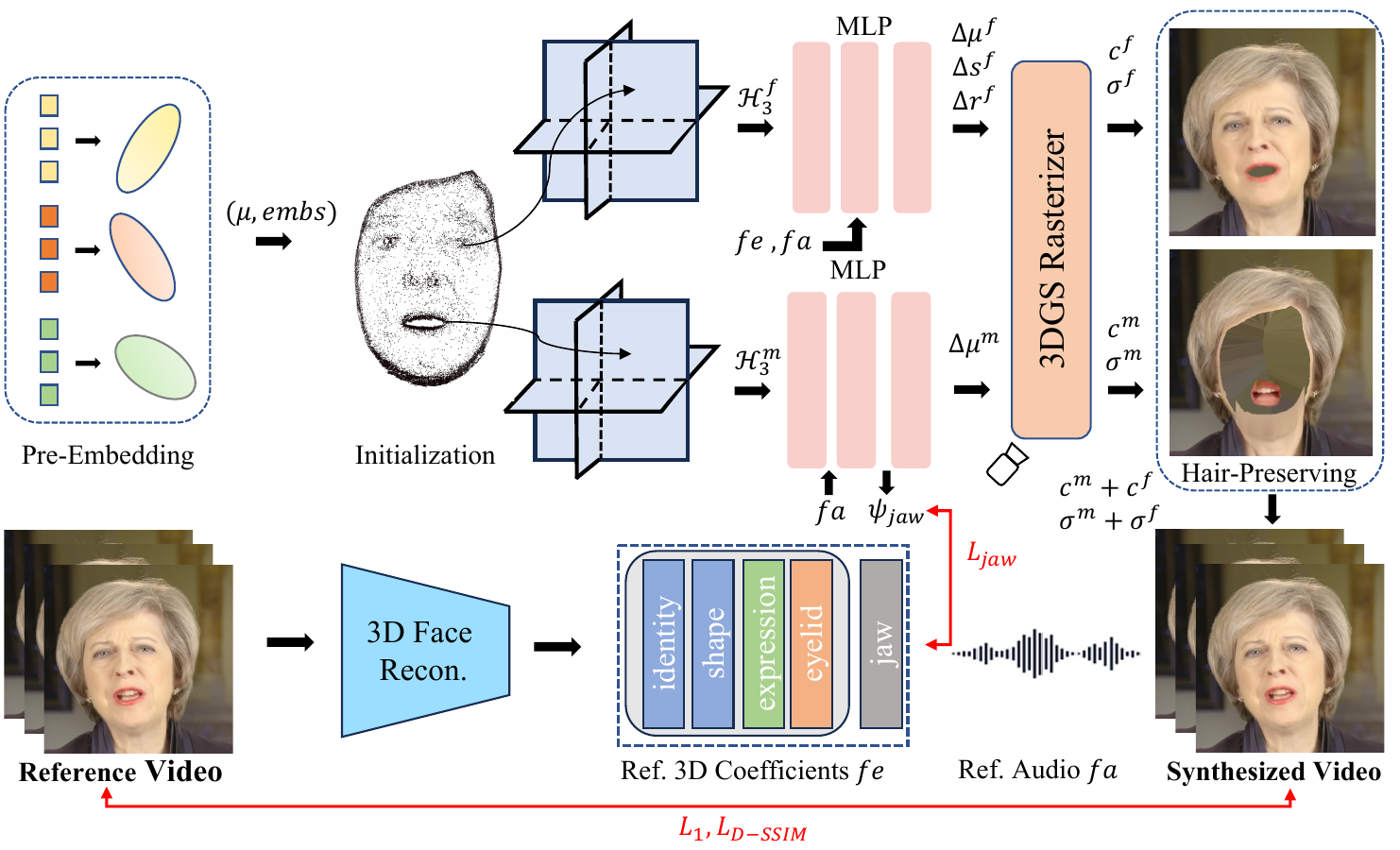}
  \caption{\textbf{Overview of DEGSTalk. } Given a cropped reference video of a talking face and its corresponding speech, our DEGSTalk first extracts the audio feature $f_a$ and performs 3D face reconstruction \cite{retsinas20243d} to obtain 3DMM coefficients, including identity, shape, expression, eyelid movement, and jaw. Secondly, Gaussian primitives are pre-embedded to construct the deformable pre-emebdding Gaussian fields and then optimize the coarse static fields of the face and mouth from random point clouds. These deformable Gaussian fields predict transformations in position, scale, and rotation. Then, the 3DGS rasterizer deforms and renders the pre-embedding Gaussian primitives into 2D images from the camera perspective for face and mouth. Finally, these regions are fused to synthesize the final talking face video using a dynamic hair-preserving portrait rendering.}

  \label{fig:overview}
\end{figure*}

Talking face synthesis methods can be divided into 2D-based and 3D-based approaches. 2D-based methods, often rely on Generative Adversarial Networks (GANs)\cite{goodfellow2014generative}, image-to-image translation techniques \cite{isola2018imagetoimagetranslationconditionaladversarial}, and Diffusion Models\cite{rombach2021highresolution}. These methods \cite{wav2lip, zhang2023sadtalker,iplap_CVPR, ZhouMakeltTalk_2020,shen2023difftalk, shen2024diffclipleveragingstablediffusion, xu2024hallo} often focus on mapping audio to lip movements, achieving good lip synchronization but sometimes neglecting accurate head poses and facial expressions. In contrast, 3D-based approaches, particularly those using Neural Radiance Fields (NeRF) \cite{mildenhall2020nerf}, model continuous 3D scenes
using implicit functions. However, they are computationally demanding \cite{guo2021ad}. Recent advancements \cite{tang2022rad, li2023ernerf, peng2023synctalk,ye2023geneface, ye2023geneface++} have improved their efficiency for real-time application. Particularly, a significant recent development is 3D Gaussian Splatting (3DGS) \cite{kerbl3Dgaussians}. Unlike NeRF, which utilizes implicit volumetric representations, 3DGS uses explicit Gaussian models. These 3DGS-based approachs\cite{li2024talkinggaussian, cho2024gaussiantalker,yu2024gaussiantalkerspeakerspecifictalkinghead, chen2024gstalker} for talking face synthesis enhances both training speed and rendering efficiency, achieving high frame rates over 100 FPS with short training times, while maintaining high-quality synthesis.

Despite these advances, current methods still face significant challenges. One major problem is distortions in dynamic facial regions. NeRF-based methods\cite{guo2021ad,tang2022rad,li2023ernerf,peng2023synctalk} struggle to accurately capture the complex features of these regions with their continuous and smooth neural fields. Similarly, 3DGS-based methods\cite{li2024talkinggaussian, cho2024gaussiantalker,yu2024gaussiantalkerspeakerspecifictalkinghead, chen2024gstalker}, which rely only on 3D Gaussians, frequently fail to accurately represent all facial features, especially in individuals with more complex facial dynamics. And these methods with explicit expression representations based on facial landmarks \cite{ye2023geneface, ye2023geneface++} or action units (AU) \cite{li2023ernerf, peng2023synctalk, li2024talkinggaussian}, which limits their ability to fully capture intricate facial dynamics. Additionally, reconstructing individuals with long hair remains challenging, as the complex interactions between hair and the head-torso connection often introduce noise or artifacts into the generated videos. For example, recent methods \cite{peng2023synctalk} that attempt to address these issues, frequently result in new artifacts, such as distortions of double chin.

In this paper, we propose DEGSTalk, a 3D Gaussian Splatting (3DGS)-based method utilizing Deformable Pre-Embedding Gaussian Fields for hair-preserving talking face synthesis. At first, we propose the Deformable Pre-Embedding Gaussian Fields, which enhances the realism and adaptability of synthesized faces by dynamically adjusting pre-embedding Gaussian primitives with enhanced implicit expression coefficients, to capture the fine features of facial regions and expressions. This method allows for a more accurate representation of dynamic facial regions, overcoming the distortions typically observed in previous approaches. Second, we develop a Dynamic Hair-Preserving Portrait Rendering technique for hair-preserving talking face synthesis, which is specifically designed to tackle the challenge of accurately reconstructing individuals with long hair.


\section{PROPOSED METHOD}
\label{sec:method}

\subsection{Preliminaries and Problem Setting}
\label{ssec:preliminaries}

\vspace{0pt}\noindent\textbf{3D Gaussian Splatting.} 
3DGS \cite{kerbl3Dgaussians} represents 3D information using a set of 3D Gaussian. Each Gaussian primitive is defined by parameters including a center $\mu \in \mathbb{R}^3$, a scaling factor $s \in \mathbb{R}^3$, a rotation quaternion $q \in \mathbb{R}^4$, an opacity value $\alpha \in \mathbb{R}$, and a color feature $f \in \mathbb{R}^Z$. The set of parameters for each Gaussian can be denoted as $\theta_i = \{\mu_i, s_i, q_i, \alpha_i, f_i\}$.
\begin{equation}
    \setlength{\abovedisplayskip}{4pt}
    \setlength{\belowdisplayskip}{4pt}
    \mathcal{G}_i(\mathbf{x}) = e^{-\frac{1}{2}(\mathbf{x-\mu_i})^T\Sigma_i^{-1}(\mathbf{x-\mu_i})},
\end{equation}
To facilitate optimization, the covariance matrix $\Sigma$ can be calculated from $s$ and $q$.

During the rendering process, the color $\mathcal{C}$ is computed by blending $\mathcal{N}$ ordered points overlapping the pixel, as described by the following equation:
\begin{equation}
    C = \sum_{i \in \mathcal{N}} c_i \alpha_i \prod_{j=1}^{i-1} (1 - \alpha_j), \qquad  \quad \alpha_i = \sigma_i\mathcal{G}_i(\mathbf{x}),
    \label{eq:blending}
\end{equation}
where $c_i$ is the color of each point decoded with color feature $f$, and $ {\alpha}_i^\prime$ is computed by the multiplication of the opacity $\alpha$ of the 3D Gaussian and its projected Gaussian $\mathcal{G}_i(\mathbf{x})$. 

This approach optimizes the Gaussian parameters $\theta$ through gradient descent under color supervision, with strategies like densification and pruning employed to manage the number of active primitives during the rendering process.

\vspace{0pt}\noindent\textbf{Problem Setting.} 
Similar to previous NeRF-based approaches \cite{guo2021ad,liu2022semantic,tang2022rad,li2023efficient}, 
TalkingGaussian \cite{li2024talkinggaussian} introduces a 3DGS-based audio-driven framework for high-fidelity talking face synthesis. Based on its setting, we aim to predict a point-wise deformation $\delta_i = \{\Delta 
\mu_i, \Delta s_i, \Delta q_i\}$ for each primitive with the input of its center $\mu_i$, audio feature $f_{a}$ and expression $f_{e}$ and an efficient and expressive
tri-plane hash encoder $\mathcal{H}$ for position encoding with an MLP decoder.
The deformation $\delta_i$ for the $i$-th primitive in the face branch can be predicted by: 
\begin{equation}
    \setlength{\abovedisplayskip}{4pt}
    \setlength{\belowdisplayskip}{4pt}
    \delta_i = \text{MLP}(\mathcal{H}(\mu_i) \oplus f_{a} \oplus f_{e}),
\end{equation}
where $\oplus$ denotes concatenation.

Through a rasterizer in 3DGS \cite{kerbl3Dgaussians}, the fields are combined to generate deformed Gaussian primitives to render the output image, of which the deformed parameters $\theta_D$ are calculated from the canonical parameters $\theta_C$ and deformation $\delta$:
\begin{equation}
    \setlength{\abovedisplayskip}{4pt}
    \setlength{\belowdisplayskip}{4pt}
    \label{eq:deform_param}
    \theta_D = \{\mu+\Delta \mu, s+\Delta s, q+\Delta q, \alpha, f\}.
\end{equation}

\subsection{Deformable Pre-Embedding Gaussian Fields with Enhanced Face Expression}
\label{ssec:preembedding}

As shown in Fig \ref{fig:overview}, different from traditional methods that only rely on Gaussians, our DEGSTalk introduces a dynamic embedding learning strategy where each Gaussian primitive is associated with a learnable embedding $\emb \in \mathbb{R}^d$. In Deformable Pre-Embedding Gaussian Fields, these embeddings encode nuanced information related to facial regions, capturing subtle expressions and dynamic movements more effectively. By dynamically adjusting the Gaussian primitives through these embeddings, our DEGSTalk can precisely model complex facial deformations and expressions, improving realism and adaptability in talking face synthesis, especially in challenging scenarios with diverse facial features and movements.

Additionally, inspired by recent advanced single-image deep 3D reconstruction methods \cite{retsinas20243d}, we employ predicted implicit 3D Morphable Model (3DMM) coefficients as our intermediate facial representation. These implicit 3DMM coefficients—including identity $\psi_{id}$, shape $\psi_{s}$, expression $\psi_{exp}$, jaw $\psi_{jaw}$, and eyelid movements $\psi_{eye}$—offer a more flexible and comprehensive way to encode facial geometry and motion compared to explicit representations, which usually require manual adjustments and struggle with complex dynamics. By leveraging the implicit 3DMM, our approach achieves a more natural and detailed reconstruction of facial features. We use these coefficients, denoted as $f_{e} = \{\psi_{id}, \psi_{s}, \psi_{exp}, \psi_{eye}\}$, as facial features. The facial features, along with the audio input, guide the Deformable Pre-Embedding Gaussian Fields to generate talking faces.

Specifically, we propose an efficient tri-plane hash encoder \(\mathcal{H}\) with position encoding and an MLP decoder (similar to \cite{li2024talkinggaussian}) to ensure smooth and continuous facial motion. The deformation fields represents facial motion by mapping the center \(\mu\), the canonical Gaussian parameters \(\theta_C = \{\mu, s, q, \alpha, f\}\), and the Gaussian embedding \(\emb\) to the deformation parameters as:
\begin{equation}
\begin{aligned}
\delta_D &= \text{MLP}(\mathcal{H}(\mu \oplus \emb) \oplus f_{a} \oplus f_{e}) \\ &= \{\Delta \mu, \Delta s, \Delta q\},
\label{eq:deformed_parameters}
\end{aligned}
\end{equation}
The transforms in position, scale, and rotation can be represented as :
\begin{equation}
\begin{aligned}
    \theta_D &= \theta_C + \delta_D \\
             &= \{\mu + \Delta \mu, s + \Delta s, q + \Delta q\},
\end{aligned}
\label{eq:deformed_function}
\end{equation}
where \(\theta_D\) represents the deformed Gaussian parameters for rendering frames and \(\oplus\) denotes feature concatenation.




\subsection{Dynamic Hair-Preserving Portrait Rendering}
\label{ssec:render}

To address the challenge of accurately reconstructing individuals with long hair in talking face synthesis, we propose a novel method called Dynamic Hair-Preserving Portrait Rendering. While traditional approaches \cite{guo2021ad, tang2022rad,li2023ernerf,li2024talkinggaussian} usually struggle to maintain the natural flow and appearance of hair, 
our rendering effectively overcomes these issues by spatially integrating hair and face regions and preserving their distinct characteristics.

Similar to the approach used in TalkingGaussian \cite{li2024talkinggaussian}, our methods start with separately reconstructing the face and mouth regions, explicitly excluding the hair. 
However, the direct fusion of the face and hair regions will introduce significant noise and artifacts. To address it, we employ Dynamic Hair-Preserving Portrait Rendering, which carefully integrates the entire face region with the preserved hair. Initially, the face and mouth regions undergo dilate operations during training to better capture their features, while also preserving the hair region $\mathcal{C}_\text{hair}$. 
Guided by the densities ($\sigma^f$) and colors ($\mathbf{c^f}$) of the face region, and the densities ($\sigma^m$) and colors ($\mathbf{c^m}$) of the mouth region from deformable gaussian fields.
Then we fine-tune deformable Gaussian fields to fuse the face and hair region, which is subsequently subjected to two additional rounds of rendering. 
Consequently, the talking face color $\mathcal{C}$ can be rendered by:
\begin{equation}
\label{eq:fusion}
    \mathcal{C} = \mathcal{C}_\text{hair} + \mathcal{C}_\text{face} \times \mathcal{O}_\text{face} + \mathcal{C}_\text{mouth} \times (1 - \mathcal{O}_\text{face}),
\end{equation}
where  $\mathcal{C}_\text{face}$ and $\mathcal{O}_\text{face}$ denote the predicted face color and opacity of the face region, and $\mathcal{C}_\text{mouth}$ is the mouth region color.

Dynamic Hair-Preserving Portrait Rendering technique ensures that the resulting talking face animations are both realistic and cohesive. By integrating hair and facial features in a way that maintains their individual properties, we achieve high-quality renderings. This highlights the effectiveness of our approach in preserving hair dynamics while enhancing the overall realism of facial expressions in audio-driven talking face synthesis.


\subsection{Training Details}
\label{ssec:training}

The training process of DEGSTalk follows the three-stage pipeline of TalkingGaussian \cite{li2024talkinggaussian}: Static Initialization, Motion Learning, and Fine-Tuning. In the \textbf{Static Initialization phase}, we optimize the canonical Gaussian parameters $\theta_C$ using pixel-wise L1 loss and D-SSIM to construct a coarse head structure. 
Then, the \textbf{Motion Learning stage} applies a jaw coefficient loss to optimize the deformation parameters $\theta_D$, focusing on smooth facial dynamics and accurate lip synchronization. The loss function is computed using L1, D-SSIM, and a jaw movement loss between the masked ground-truth image $\mathcal{I}_\text{mask}$ for each region and the rendered image $\mathcal{I}_D$:
\begin{equation}
    \mathcal{L}_D = \mathcal{L}_1(\mathcal{I}_D, \mathcal{I}_\text{mask}) + \lambda \mathcal{L}_{\mathrm{D-SSIM}}(\mathcal{I}_D, \mathcal{I}_\text{mask}) + \beta \mathcal{L}_{\mathrm{jaw}},
    \label{eq:motion_loss_jaw}
\end{equation}



Finally, the \textbf{Fine-Tuning stage} refines the full face using L1, D-SSIM\cite{baker2022dssim}, and LPIPS losses to ensure realistic rendering between the fused image $\mathcal{I}_\text{fuse}$ and the ground-truth video frame $\mathcal{I}$. At this stage, only the color parameters are updated to prevent overfitting.
\begin{equation}
    \setlength{\abovedisplayskip}{5pt}
    \setlength{\belowdisplayskip}{5pt}
    \mathcal{L}_F = \mathcal{L}_1(\mathcal{I}_\text{fuse}, \mathcal{I}) + \lambda \mathcal{L}_{\mathrm{D-SSIM}}(\mathcal{I}_\text{fuse}, \mathcal{I}) + \gamma \mathcal{L}_{\mathrm{LPIPS}}(\mathcal{I}_\text{fuse}, \mathcal{I}).
\end{equation}

\vspace{-0.12cm}
\section{EXPERIMENTS AND RESULTS ANALYSIS}
\label{sec:experiment}
\subsection{Experimental Settings}
\noindent\textbf{Dataset. } 
For our experiments, we use publicly available video datasets from previous works \cite{tang2022rad, ye2023geneface, audio-driven-talking-head-generation}, which include three male portraits (``Macron", ``Obama", ``Jae-in") and three female portraits (``May", ``Eng1", ``Shaheen"). Our dataset consists of six high-definition video clips of talking faces, each with an average length of approximately 7,000 frames recorded at 25 frames per second (FPS). These raw videos are cropped and resized to a resolution of 512 $\times$ 512, with the exception of Obama, which is resized to 450 $\times$ 450 to match its original format. 

\noindent\textbf{Implementation Details.}
We follow SyncTalk \cite{peng2023synctalk} for estimating pose parameters. We first train the face and mouth regions for 50,000 iterations, followed by fine-tuning for an additional 10,000 iterations. We use the Adam \cite{kingma2014adam} and AdamW\cite{loshchilov2018adamw} optimizers across all modules, with $\gamma$, $\lambda$ and $\beta$ in the loss functions set to 0.2, 0.5 and 0.001, respectively. The dimension $d$ of embedding $\emb$ is set to 32. 

\noindent\textbf{Evaluate Metrics. } 
We evaluated the performance of the models using several established metrics to ensure accurate and comprehensive assessment. These included Peak Signal-to-Noise Ratio (PSNR) to measure the image quality, Learned Perceptual Image Patch Similarity (LPIPS) \cite{zhang2018unreasonable} for perceptual similarity, Landmark Distance (LMD) \cite{chen2018lmd} for evaluating the accuracy of facial features, and Structural Similarity Index (SSIM) \cite{wang2004ssim} to quantify the realism of the generated images.

\vspace{-0.2cm}
\subsection{Quantitative Results}
We compared our approach with several NeRF-based models, including AD-NeRF \cite{guo2021ad}, ER-NeRF \cite{li2023ernerf}, and SyncTalk \cite{peng2023synctalk}. Additionally, we included the 3D Gaussian Splatting-based model, TalkingGaussian\cite{li2024talkinggaussian}, as a baseline for high-speed performance. As shown in Table \ref{tab:setting1}, our image quality is superior to other methods in all aspects. The inference speed of our method reaches 114 FPS, which is slightly lower than the 120 FPS of TalkingGaussian\cite{li2024talkinggaussian}. In comparison, our DEGSTalk achieves the best performance in all four metrics, which demonstrates its high efficiency.

\vspace{-0.2cm}
\begin{table}[h]
\begin{flushleft}  
\resizebox{1\linewidth}{!}{  
        \setlength{\tabcolsep}{2.5mm}  
        
        \begin{tabular}{lccccccccc}
        \toprule
        Methods & PSNR $\uparrow$ & LPIPS $\downarrow$ & LMD $\downarrow$ &  SSIM $\uparrow$ & Time & FPS      \\ \midrule
        AD-NeRF \mysmall{(ICCV 21 \cite{guo2021ad})}       & 29.18     & 0.0963        & 2.888         &   0.7670    & 18h   & 0.16 \\
        ER-NeRF \mysmall{(ICCV 23 \cite{li2023ernerf})}& 32.06     & 0.0354        &  2.771        &   0.9158    & 2h    & 45   \\ 
        SyncTalk  \mysmall{(CVPR 24 \cite{peng2023synctalk})}    & 33.73     & 0.0292        &  2.627        &    0.9361   & 2h    & 60  \\ 
        TalkingGaussian \mysmall{(ECCV 24 \cite{li2024talkinggaussian})} & 32.39  & 0.0369 & 2.659 & 0.9261       &  \textbf{0.5} & \textbf{120}       \\ 
        Our & \textbf{37.30}   & \textbf{0.0140} & \textbf{2.349} & \textbf{0.9664} &  \textbf{0.5} & 114  \\
        \bottomrule 
        
        \end{tabular}
    }
    \setlength{\abovecaptionskip}{0cm}
    
    \vspace{0em} 
    \caption{The quantitative results of the \emph{head reconstruction setting}. The best  methods are in \textbf{bold}.}
    \label{tab:setting1}
\end{flushleft}  
\end{table}

\vspace{-1cm}
\subsection{Qualitative Evaluation}
\begin{figure}[t]
    
    \includegraphics[width=1\linewidth]{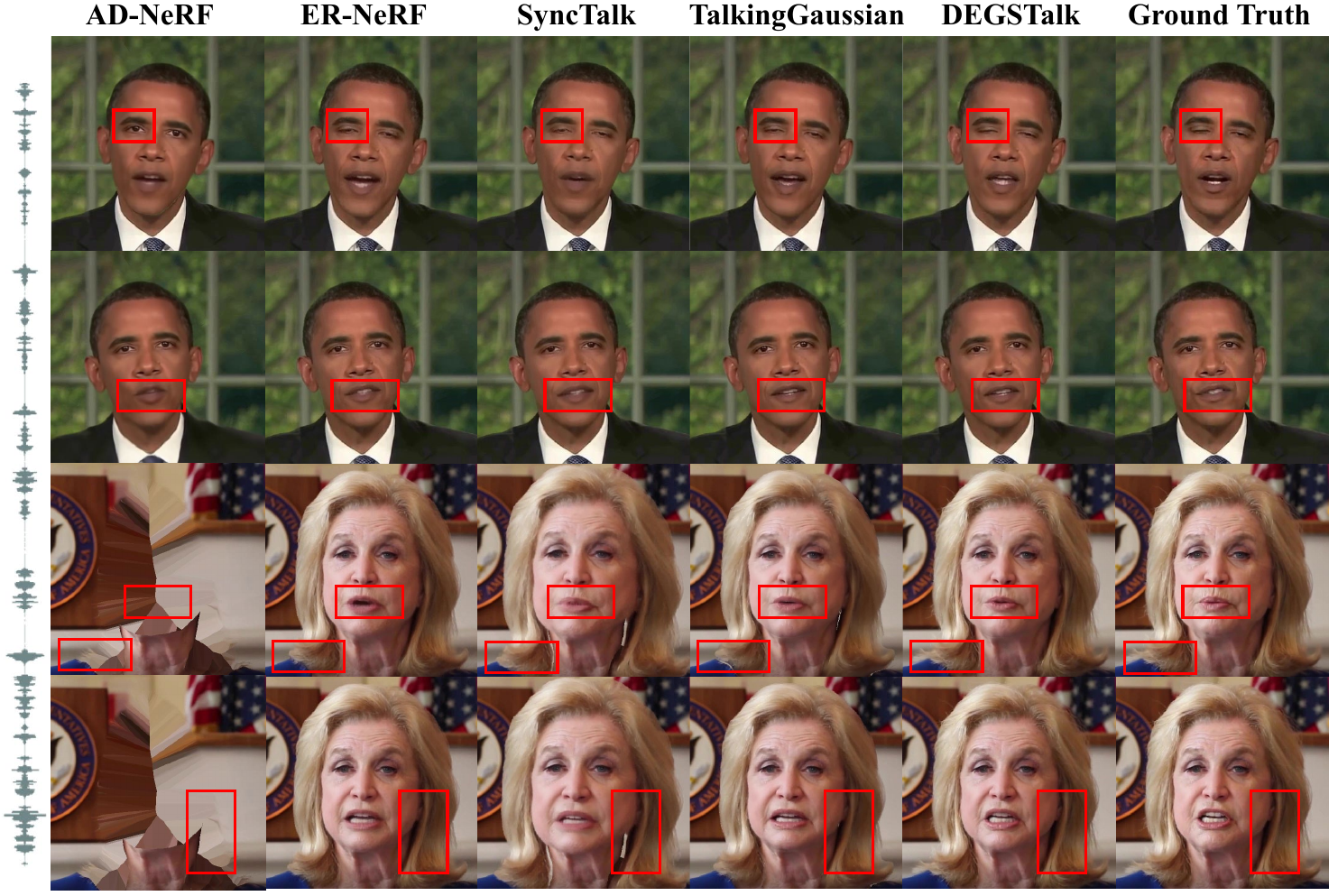}
    
    \caption{Visual results of the comparative experiments. We show the generated results of the baselines under the head reconstruction setting and the ground truth.}
    \label{fig:exp1}
    \vspace{-0.55cm}
\end{figure}

To qualitatively evaluate the synthesis quality, we show keyframes of a reconstructed sequence from the head reconstruction setting and details of two portraits in Fig.~\ref{fig:exp1}. From the figure, our method shows significant improvements in the reconstruction of long-haired individuals compared to baseline methods while AD-NeRF struggles to reconstruct long-haired individuals. In particular, AD-NeRF struggles with long hair, while DEGSTalk preserves fine hair details and avoids artifacts commonly seen in other methods (ER-NeRF, SyncTalk, and TalkingGaussian). Additionally, DEGSTalk leads to more accurate and reliable facial expressions, particularly in dynamic regions like the eyes and mouth. The consistency and precision in both hair and facial expressions demonstrate the robustness of our method in handling complex talking face scenarios. Whether compared to NeRF-based or 3DGS-based methods, DEGSTalk consistently outperforms them.

\vspace{-0.2cm}
\subsection{Ablation Study}  
\begin{figure}[h]

\centering    \includegraphics[width=0.9\linewidth]{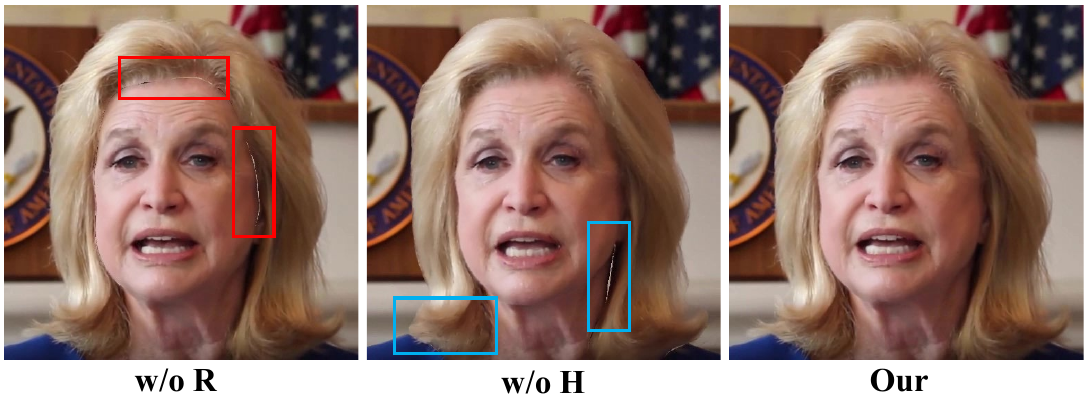}
    \setlength{\abovecaptionskip}{-0.1cm}
    \caption{Visual results of the ablation Study.}
    \label{fig:exp2}
    \vspace{-0.3cm}
\end{figure}

We conducted ablation studies to assess the contribution of key components in our framework: the pre-embedding process (E), the use of implicit facial coefficients (F), direct fusion during rendering (R), and hair-preserving rendering (H).  
As shown in Fig. \ref{fig:exp2}, R and H address issues related to the reconstruction of long hair, particularly in reducing artifacts at the junction between the face and hair (red regions) and at the connection between the neck and head (blue regions).
Furthermore, Table \ref{tab:ablation} shows that our method achieves strong performance across multiple evaluation metrics.

\vspace{-1.2em}
\begin{table}[h]
\begin{flushleft}  

\vspace{-0.3cm}
\resizebox{1\linewidth}{!}{  
        \setlength{\tabcolsep}{2.5mm}  
        
        \begin{tabular}{lccccccccc}
        \toprule
        Methods & PSNR $\uparrow$ & LPIPS $\downarrow$ & LMD $\downarrow$  & SSIM $\uparrow$      \\ \midrule
        w/o F & 36.79 & 0.150 & 2.630 & 0.9692\\  
        w/o R & 35.66& 0.0300& 2.359& 0.9678&\\ 
        w/o E & 37.21 & 0.0141& 2.357 & \textbf{0.9712} \\
        w/o H & 33.89 & 0.0301& 2.628 & 0.9354\\
        w E, F, R, H  & \textbf{37.30}  & \textbf{0.0140} & \textbf{2.349} & 0.9664      \\ 
        \bottomrule 
    
        \end{tabular}
    }
    \setlength{\abovecaptionskip}{0cm}
    
    \vspace{0em} 
    \caption{The quantitative results of ablation Study.}
    \label{tab:ablation}
\end{flushleft}  
\end{table}

\vspace{-0.8cm}
\section{CONCLUSIONS}
\label{sec:conlusions}

In this paper, we propose a novel Deformable Pre-Embedding Gaussian Fields (DEGSTalk) enhancing the realism and adaptability for hair-preserving talking face synthesis. It addresses the challenges of reconstructing individuals with long hair to perform in more accurate and lifelike talking face synthesis. 
Results show that our method achieved state-of-the-art performance across key metrics, while largely improving training speed and rendering efficiency. Some noisy primitives may still occur. This issue will be addressed in future work.

\bibliographystyle{IEEEbib}
\bibliography{strings,refs}


\end{document}